\documentclass{article}
\usepackage{arxiv}
\usepackage{amsmath,amssymb,amsfonts}
\usepackage{algorithmic}
\usepackage{graphicx}
\usepackage{textcomp}
\usepackage{xcolor}
\usepackage{hyperref}

\usepackage[sorting=none,backend=biber]{biblatex}
\usepackage{twemojis}
\usepackage{adjustbox}
\begin{document}

\title{Hey Pentti, We Did (More of) It!: \\ A Vector-Symbolic Lisp With Residue Arithmetic
}

\author{Connor Hanley, Eilene Tomkins-Flanagan, Mary Alexandria Kelly\\ANIMUS Lab, Carleton University
}

\maketitle

\begin{abstract}
Using Frequency-domain Holographic Reduced Representations (FHRRs), we extend a Vector-Symbolic Architecture (VSA) encoding of Lisp 1.5 with primitives for arithmetic operations using Residue Hyperdimensional Computing (RHC).  Encoding a Turing-complete syntax over a high-dimensional vector space increases the expressivity of neural network states, enabling network states to contain arbitrarily structured representations that are inherently interpretable. We discuss the potential applications of the VSA encoding in machine learning tasks, as well as the importance of encoding structured representations and designing neural networks whose behavior is sensitive to the structure of their representations in virtue of attaining more general intelligent agents than exist at present.
\end{abstract}

\section{Introduction}

\textcite{Flanagan2024} define an encoding for Lisp \cite{McCarthy1960} in vector-space, as well as an interpreter defined over the
encoded representations. This allows for the implementation of any computable procedure, and for the computation of that procedure,
entirely in vector-space. To do this, they encode Lisp expressions in a vector-symbolic architecture, which when paired with a memory system, 
is as expressive as a Turing machine. In this paper, we contribute a novel integer encoding using Residue Hyperdimensional Computing\footnote{Our implementation may be found at \url{https://github.com/hanleyc01/residuelisp}}.
Residue Hyperdimensional Computing is a vector-symbolic architecture based on residue arithmetic, allowing for a unique code for any integer, as
well carry-free arithmetic operations. In \autoref{formal-vsa}, we recapitulate the generalization of vector-symbolic algebras used in \textcite{Flanagan2024}.
In \autoref{rhc}, we give a short introduction to Residue arithmetic and the Residue Hyperdimensional Encoding proposed by \textcite{kymn_computing_2024}. In \autoref{resonator}, we define Resonator
networks following \textcite{kent_resonator_2020, frady_resonator_2020}, a state-of-the-art method for factorizing complex vector-symbolic expressions, used in the implementation
of Residue Hyperdimensional Computing. In \autoref{vsa-lisp} we discuss our novel contributions to the the encoding of Lisp in vector-symbolic architectures.
In Section~\ref{encoding}, we provide definitions of the encoding of integers in the Lisp implementation, and associated semantic functions.
In Section~\ref{discussion}, we discuss the importance of our contributions, motivating it by providing potential concrete use cases in learning tasks, as well as by appealing to recent literature on the need for algebraic representations for general intelligence.

\subsection{Formalizing VSAs}\label{formal-vsa}

\textcite{Flanagan2024} define a \emph{generic} vector-symbolic architecture (VSA) as follows:
\begin{quote}
A vector-symbolic architecture is an algebra (i.e., a vector space with a bilinear product),
\begin{enumerate}
    \item that is closed under the product $\otimes: V \times V \to V$ (i.e., if $u \otimes v = w$, then $u, v, w \in V$)
    \item whose product has an ``approximate inverse'' $\overline{\otimes}$ that, given a product $w$ and one of its operands $u$ or $v$, yields a vector correlated with the other operand
    \item for which there is a dogma for selecting vectors from the space to be treated as atomic ``symbols'' (yielding themselves, thereby, to syntactic manipulations defined in terms of the algebra), 
    \item that is paired with a memory system $\mathcal{M}$ that stores an inventory of known symbols for retrieval after lossy operations (e.g., involution), that can be recalled from $\mathcal{M}(p)$, and which is appendable $\mathcal{M} \twoheadleftarrow t$, and
    \item possesses a measure of the correlation (a.k.a., similarity) of two vectors, $\mathbf{sim}(u, v) \in [-1, 1]$, where $1$ and $-1$ imply that $u, v$ are colinear, $0$ that they are linearly independent.
\end{enumerate}
\end{quote}

Some VSAs relax some of the preceding properties \cite{Plate1997,Schlegel2022,Kelly2011,Kleyko2022}, but have behavior approximating all listed properties. Importantly, the ``dogma for selecting symbols'' may be freely chosen, so they can be \textit{arbitrary semantic embeddings}, including the token embeddings of a neural network, photographs \cite{Kelly2013}, or (recursively) complex symbolic expressions generated by another VSA  \cite{Kelly2020indirect}. Most of our symbols will be drawn from a random overcomplete basis to minimize any possible influence of the distribution of the symbols on computations, but our numeric representations, importantly, are related to one another by a group structure \cite{Ozen2024}. Under the above definitions, it can be shown \cite{Flanagan2024} that VSAs are Cartesian closed \cite{Cartesian2024}, implying that Turing-complete languages can be described over any VSA. The VSA Lisp is accordingly described for a generic VSA, meaning that it can be implemented in any VSA. The reference implementation uses the \textit{Holographic Reduced Representations} (HRRs) described by \textcite{Plate1995}.

For this implementation, we used \textit{Fourier-domain Holographic
Reduced Representations} \cite{Plate2003, Ozen2024, yeung_generalized_2024, Kleyko2022} (FHRRs),
which are closed under the same algebraic properties, but have the added benefit of 
being a generalization over the arithmetical encoding. For example, the product $\otimes$ is 
element-wise multiplication (i.e., the Hadamard product) $\odot$ , the sum is element-wise addition $+$, the inverse of the product of two vectors $u, v$ is $(u \odot v) \odot \overline{u} \approx v$ w, and the similarity kernel for two vectors $u, v$ of dimension $D$ is
\begin{equation}
    \frac{\mathfrak{R}\{\overline{u^T} v\}}{D},
\end{equation}
with $\overline{u^T}$ being the complex conjugate transpose of $u$, and $\mathfrak{R}$ the real component of a complex vector. 
What distinguishes FHRRs from HRRs is,
instead of converting the vectors into the Fourier-domain to implement circular convolution, which is the binding
operation for HRRs, we rather remain entirely in the Fourier-domain. For a full comparison of HRRs and FHRRs, as well as 
the later introduced Residue Hyperdimensional Computing VSAs, see Table~\ref{fig1}.

\begin{table*}
\centering
\caption{Comparison of HRRs, FHRRs, and RHCs; first two reproduced from \textcite{Kleyko2022}}
\begin{adjustbox}{width=1\textwidth}
\begin{tabular}{llllll}
\hline
VSA & Space of atomic vectors & Binding & Unbinding & Superposition & Similarity \\ \hline
HRR & unit vectors & circular convolution & circular correlation & \begin{tabular}[c]{@{}l@{}}element-wise\\ additon\end{tabular} & \begin{tabular}[c]{@{}l@{}}dot product\\ similarity\end{tabular} \\
FHRR & complex unitary vectors & \begin{tabular}[c]{@{}l@{}}element-wise product/\\ Hadamard product\end{tabular} & element-wise product with inverse & \begin{tabular}[c]{@{}l@{}}element-wise\\ addition\end{tabular} & \begin{tabular}[c]{@{}l@{}}cosine\\ similarity\end{tabular} \\
RHC & subset of complex unitary vectors & \begin{tabular}[c]{@{}l@{}}additive binding and \\ multiplicative binding\end{tabular} & \begin{tabular}[c]{@{}l@{}}element-wise product with inverse/\\ multiplicative binding with inverse\\ modulo $m$\end{tabular} & \begin{tabular}[c]{@{}l@{}}element-wise\\ addition\end{tabular} & cosine similarity \\ \hline
\end{tabular}
\end{adjustbox}
\end{table*}\label{fig1}

\subsection{Residue Arithmetic and Residue Hyperdimensional Computing}\label{rhc}

A residue number system, or residue coding, is a scheme by which we can uniquely encode integers
into vectors of their value moduli a sequence of relative co-prime values \cite{1garner_residue_1959}. Let $M_n = [m_1, m_2, \ldots, m_n]$ be a vector of co-prime positive integers. To encode an arbitrary integer $x$ into a residue number system with moduli $M_n$, we would have,
\begin{equation}
    [x \bmod m_1, x \bmod m_2, \ldots, x \bmod m_n].
    \label{residue_arithmetic}
\end{equation}
We are also able to define carry-free arithmetical operations over residue numbers,
which are just element-wise sum and multiplication modulo $m_i$, for $i = 1, 2, \ldots, n$.

In the same way that FHRRs sample uniformly from the range $(0, 2\pi]$ \cite{Kleyko2022} in order
to create complex-valued elements, for some moduli $m_i$ we can sample from the $m_i$-th roots
of unity of the unit circle for vector elements, which is called Residue Hyperdimensional 
Computing \cite{kymn_computing_2024} (RHC). 
Encoding integers in $D$-dimensional RHC is a two-step process:

First, define a set of $D$ co-prime positive integer moduli 
\begin{equation}
    M_n = [m_1, m_2, \ldots, m_n].
\end{equation}
For each $m_i$, $i = 1, 2, \ldots, n$, let
\begin{equation}
    \phi_i = [\theta_1, \theta_2, \ldots, \theta_D],
\end{equation} 
where each $\theta_j$, $j = 1, 2, \ldots, D$ is a randomly sampled angle from the $m_i$-th roots of unity. 
E.g., suppose some $m_k$ were $2$. Then, each of angle in $\phi_k$ would be sampled from the set $\{\pi, 2\pi\}$. 
The residue moduli are finally encoded as:
\begin{equation}
    z_i = \exp(\sqrt{-1} \phi_i).\label{z_i}
\end{equation}

Secondly, for any integer $x$, for each $z_i$, $i = 1, 2, \ldots, n$, we define
the residue code, $\zeta_i : \mathbb{Z} \to \mathbb{C}^D$:
\begin{equation}
    \zeta_i (x) = z_i^x = \exp(\sqrt{-1} \phi_i x).
\end{equation}
To get the final RHC representation over the set of co-prime moduli $M_n$, we then take the element-wise product (or Hadamard product) of each
$\zeta_i (x)$, such that,
\begin{equation}
    \zeta_{M_n} (x) = \bigodot^n_{i=1} \zeta_i (x)
\end{equation}

RHC, with the addition of a memory system, forms a complete VSA. Instead of completely encoding Lisp in RHC, we instead opt to 
only have numerical items be represented in RHC. This is because, aside from the similarity kernel being different,
the set from which RHC is sampled from is a subset of FHRR. The unique property of 
RHC which sets it apart from other VSA is that it has two operations for binding: additive binding and multiplicative binding.
For two RHC-encoded integers $\zeta_{M_n} (x_1)$ and $\zeta_{M_n} (x_2)$, additive binding, which obeys the following property,
\begin{equation}
    \zeta_{M_n} (x_1) \odot \zeta_{M_n} (x_2) = \zeta_{M_n} (x_1 + x_2)
\end{equation}
is implemented by element-wise multiplication. 
The second binding, multiplicative binding, obeys the following property,
\begin{equation}
    \zeta_{M_n} (x_1) \star \zeta_{M_n} (x_2) = \zeta_{M_n} (x_1 x_2),
\end{equation}
where $\cdot \star \cdot$ denotes multiplicative binding.

Multiplicative binding is, unfortunately, complex. For an in-depth discussion, see \textcite{kymn_computing_2024}, Section
4.1.2 for concrete details. The simplest way to implement multiplicative binding is to perform element-wise exponentiation of the representation
$\zeta_{M_n} (x_1)$ by the integer value $x_2$: $\zeta_{M_n}(x_1)^{x_2}$. To do this, we can decode $x_2$ from $\zeta_{M_n} (x_2)$ either using codebook decoding,
or by factorizing the expression into its component root vectors using a resonator network (Section~\ref{resonator}), and then decoding the result in terms of the moduli of the RHC $M_n$.
\textcite{kymn_computing_2024} note that there is another method which does
not require decoding $x_2$ from $\zeta_{M_N} (x_2)$, but it requires decomposing $\zeta_{M_n} (x_1)$ and $\zeta_{M_n} (x_2)$ into
their component factors, exploiting the fact that each element in $\zeta_i(x)$, for some integer $x$, are angles. Implementing this method is left as a future optimization.

\subsection{Resonator Networks}\label{resonator}

A Resonator network is a system of auto-associative neural networks. It is employed 
as a memory system for factorizing composite VSA representations 
\cite{frady_resonator_2020, kent_resonator_2020}. Define the
sets of atomic symbols $X = [x_1, x_2, \ldots, x_N]$, $Y = [y_1, y_2, \ldots, y_N]$,
and $Z = [y_1, y_2, \ldots, y_N]$. Given an representation $s = x_i \odot y_j \odot z_r$,
where $x_i$, $y_j$, and $z_r$ specifically are unknown, we can estimate their values:

The update rules for the each estimated representation is given by the application of the resonator to the prior state, as follows,
\begin{equation}
    \hat x (t) = \begin{cases}
&\sum^n_{i = 1} x_i, \text{ if } t = 0 \\
&g (X X^T (s \odot \hat y (t - 1) \odot \hat z (t - 1))),
    \end{cases}\label{resnet_def}
\end{equation}
with $g$ being some thresholding function, and likewise for the other factors.

Resonator networks are only used here in decoding integers back into a human-readable form. Given moduli $M_n$,
we use $M_n$ as the codebook and we estimate $n$ factors. Once we have the $n$ factors of the integer representation,
we are able to map the representation back to the integers, using a similar decoding method as a residue
code.

\subsection{The VSA Lisp}\label{vsa-lisp}

For the VSA Lisp interpreter, we use the definitions in \textcite{Flanagan2024}. 
The only substantial change from the underlying representation barring the integer
encoding and addition of arithmetical operations is the format of lambda expression
representations. In the original implementation, evaluation is a two-step
recursive procedure in which lambda expressions, when first encountered, have their
raw syntax immediately returned en lieu of further evaluation, and only later evaluated
if featured in an application. One of the goals of \textcite{Flanagan2024} was to show that VSAs were Cartesian-closed \cite{Cartesian2024}, hence, they expressed even conditional evaluation in terms of the vector algebra. As this work has already been done, we have no need to repeat it, and choose to implement conditional evaluation, particularly the evaluation of lambda expressions, in a more traditional fashion, for the sake of code readability and efficiency of evaluation.

The second change from the initial interpreter is how we evaluate lambda abstraction expressions. Previously,
evaluation required two steps: one where we evaluate non-lambda expressions, and then a second step where
we check to see if we need to apply a lambda expression. We, however, chose to represent lambda expressions
using insight gained from our tuple encoding. Namely, that we can store the lambda expression as a \textit{chunk}, or a set of role-filler pairs marked with a tag \cite{Kleyko2022, gayler_vector_2003}. Function application
then involves: dereferencing the semantic pointer to the function chunk in associative memory, associating the 
parameters with provided arguments, and evaluation of the body of the function with the temporary
associative memory thus formed.

\section{Encoding Arithmetic Functions}\label{encoding}

The semantics of the VSA Lisp language remains the same as proposed by \textcite{Flanagan2024}, except for our additions below. The only new functions defined are those relating to integer encoding, integer testing, and 
binary mathematical operations. Let $M_n$ be our moduli $[m_1, m_2, \ldots, m_n]$.
Let our encoding function, for some integer $x$ be,
\begin{equation}
    \text{int}(x) = \zeta_{M_n} (x) + \texttt{int}.
\end{equation}
where \texttt{int} is a constant, an atomic symbol represented by some arbitrary vector. Further, let \texttt{t} and \texttt{f} denote the truth and false constants within Lisp. Finally, let $\vartheta$ be a similarity
maximum value, in the reference implementation set to $0.2$, which denotes the maximum amount of similarity between
vectors before we consider them indistinct. We therefore add the following functions
to the definition of the semantics. Let $u$ and $v$ be $D$-dimensional FHRR vectors,
\begin{align}
    \text{int?} (v) &= \mathcal{M}(\text{sim}(v, \texttt{int}) \texttt{t} + (2 \vartheta - \text{sim}(v, \texttt{t})) \texttt{f} ),\label{int?} \\
    \text{add}(u, v) &= ((u - \texttt{int}) \odot (v - \texttt{int})) + \texttt{int},\label{add} \\
    \text{mul}(u, v) &= ((u - \texttt{int}) \star (v - \texttt{int})) + \texttt{int},\label{mul} \\
    \text{sub}(u, v) &= \text{add}(u, \overline{v}),\label{sub} \\
    \text{div}(u, v) &= \text{mul}(u, v^{(-1 \bmod M)})\label{div}.
\end{align}
In plainer terms: (\ref{int?}) defines a type-checking operation for integers which, using the clean-up memory $\mathcal{M}$, gravitates the value
to either \texttt{t} if the value is similar to \texttt{int}, or \texttt{f} otherwise. Equation (\ref{add}) defines the addition
operation, which removes the \texttt{int} tag on both operands, and performs RHC additive binding, tagging the result as an integer.
Equation (\ref{mul}) does a similar unwrapping of the tag and performs RHC multiplicative binding. Equation (\ref{sub}) performs addition
between $u$ and the additive inverse, which is the complex conjugate of $v$. And finally, (\ref{div}) performs division
between $u$ and $v$, which is captured by multiplicative binding between $u$ and the modular multiplicative inverse of $v$.

\subsection{Complexity Analysis}

The integer representations and binding operations we use above are vastly superior in terms of efficiency relative to the reference implementation in \textcite{Flanagan2024}. In their reference code, positive integers are represented as recursive lists, where the number $x_1$ is represented by the $x_1$-nested empty list \verb|(...(nil))|, with \verb|nil| treated as $0$. Addition is a recursive operation subtracting $1$ from the operand $x_2$ (i.e., unnesting it one level), and adding $1$ to the operand $x_2$ (increasing the nesting level by one), until one operand is \verb|nil|, returning additionally $x_2$-nested \verb|(...(|$x_1\verb|))|$, an $O(mx_2) \approx O(x_2^2) \approx O(x_1^2)$ operations, where $m$ is the size of memory, as subtraction requires a retrieval. In our implementation, addition is a simple Hadamard product, so it requires no retrievals and is invariant with the values $x_1, x_2$, hence $O(1)$ as a function of $m, x_1, x_2$. \textcite{Flanagan2024} do not represent multiplication, but going by their simple implementation, it might be represented as multiple additions, where the representation of $x_1 + x_1$ is applied $x_2$ times. In which case, it would be $O(mx_1x_2) \approx O(x_2^3) \approx O(x_1^3)$. Our implementation of multiplication is bottlenecked by the residue network decoding, making it $O(n) \approx O(\log(N))$, as we have $n = O(\log(N))$, where $N$ is the largest integer we can represent, given we only need a $\log$ factor of moduli to accurately represent the positive integers $x_1, x_2$. We treat the dimension $D$ as a constant across all operations.

\begin{figure}
    \centering
    \includegraphics[width=1\linewidth]{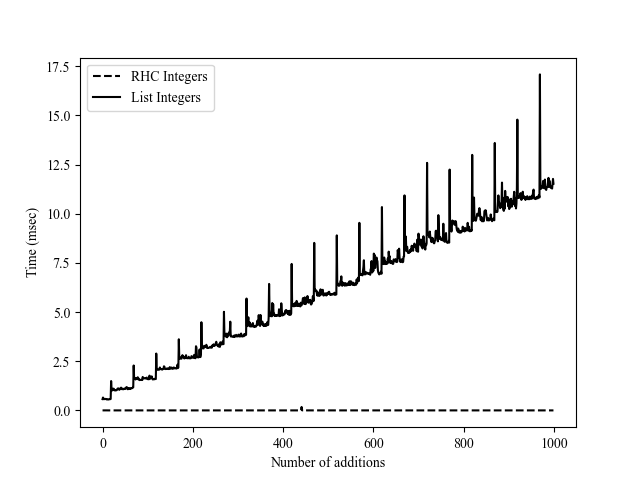}
    \caption{RHC integer encoding vs. list encoding, $D = 1000$}
    \label{fig:rhc-versus-list}
\end{figure}

To demonstrate empirically the computational superiority of the encoding, we did a rough performance test comparing iterated additions.
The results are displayed in Fig.~\ref{fig:rhc-versus-list}.
For the list encoding of the integers, we used the following function:
\begin{align}
    \mathrm{add}(x, \mathrm{nil}) &= \mathrm{nil} \\
    \mathrm{add}(x, \mathrm{cons} (\mathrm{nil}, y)) &= \mathrm{cons}(\mathrm{nil}, \mathrm{add}~(x, y)).
\end{align}
For the RHC encoding, we used RHC binding (which, recall, is component-wise multiplication between the two vectors).

\section{Discussion}\label{discussion}

In cognitive science, mental states are traditionally understood to be structured, compositional, and productive in an 
algebraic way \parencite{Fodor1988}. We suppose this to be the case as there does not seem to be, in principle, any limits to the kinds
of things we can express in thought, aside from the lack of memory capacity or computing power. The expressivity of the language of human thought imposes a lower bound on the type of machine that could account for the human mind. 

For example, one of the simplest computational machines, a finite-state automaton, can never express simple languages like those that require balanced parentheses \cite{chomsky_schutzenberger_1959}
\cite[p.~142]{Hopcroft1979}, such as Lisp, even given arbitrarily large amounts of states and transitions. In order for a finite-state automaton to approximate the behavior of a higher class of automaton, it must become exponentially more complex as a function of the maximum length of expressions to be modeled.
\textcite{Hahn2020}'s finding that Transformer Neural Networks (TNNs) trained by gradient descent and back-propagation are no more expressive than finite state automaton supplies a natural explanation for the exponential-linear tradeoff of transformer scaling laws \cite{kaplan_scaling_2020}, if learning models can be developed to take advantage of their structure.

As has been demonstrated by \textcite{Flanagan2024}, VSAs provide a natural way of expressing the combinatorial, generative properties of mental representations (as their Turing-completeness implies that they can encode the maximum class of grammar, the unrestricted grammars, corresponding to the recursively enumerable languages \cite{Hunter2021}),
while also being neurally instantiable \cite{Bekolay2014}. VSAs, therefore, may help overcome the technical limitations of present neural network paradigms.
Our contributions are two-fold:

\begin{enumerate}

\item We pair the VSA Lisp language proposed by \textcite{Flanagan2024} with state-of-the-art, efficient integer representations, expanding the expressivity of the language to domains of interest that are naturally represented through
integers. For example, suppose one is interested in modeling a task which involves indexing into some data structure 
(e.g., either an $n$-back task \cite{kirchner_age_1958}, or learning over tree positions \cite{soulos2024compositional}), then one can 
express and learn over both the indices of the data structure as well as operations defined in Lisp.

\item With respect to Transformer Neural Networks \cite{Vaswani2017}: while seeing incredible advances recently in language production,
image synthesis, and even robotics, TNNs are known to only gain expressive power as a $\log$ factor proportional to network size \cite{kaplan_scaling_2020}, and can approximate any procedure to arbitrary precision only given it may be 
\textit{arbitrarily} wide or deep \cite{Hahn2020}.  If our models can be made to have representations  with the expressivity of \textit{human} mental representations, we may be able to construct models with the 
exponentially superior scaling properties of human brains compared to TNNs:
\begin{enumerate}
    \item greater representational capacity as a function of size,
    \item lower relative energy cost as a function of task performance, and
    \item pace of skill acquisition relative to available data.
\end{enumerate}
The above properties suggest that in order to implement a general-intelligent agent \emph{that achieves its general intelligence with any
reasonable efficiency}, we should expect that we \textit{must} equip the model with similar representational
capacities to humans, which we have reason to believe are capable of expressing arbitrary programs \cite{Flanagan2024, Flanagan2025-generalintelligence, Dehaene2022, Newell1980}. Specifically, the general intelligent agent should be capable of searching through program space \cite{Chollet2019, Pennachin1998, Hutter1998} in order to find procedures that accomplish their goals.
\end{enumerate}

It is notable that Large Language Models (LLMs) have \textit{already} begun moving in the direction of greater expressivity. The limitations we listed above refer to TNNs, which we take to be typified by the Markov chain-style single token generation of the \textcite{Vaswani2017} paper, but do \textit{not necessarily} refer to the larger class of LLM architectures that generate outputs in a more complex way. For instance, \textcite{Chollet2024-Blog} writes that it appears that GPT-o3 ``searches over the space of possible Chains of Thought (CoTs) describing the steps required to solve the task, in a fashion perhaps not too dissimilar to AlphaZero-style Monte-Carlo tree search''. Chollet claims that this Monte-Carlo augmentation of o3's generations is ``natural language program search and execution within token space'', although he does not define the class of programs able to be generated by o3's methods. At least one open-source replication of OpenAI's models takes the same approach Chollet suspects of o3 in order to obtain similar results to the earlier o1 model \cite{Qin2024}, while another \cite{DeepSeek2024} (whose architecture \cite{DeepSeek2025} recently claimed parity with o1) achieves its results \textit{without} using explicit search when generating output (the authors express that their attempt at using search did not succeed, but future iterations \textit{may}). 

The reason for the discrepancy between these two approaches may be that the present state of ``deep-learning guided program search'' \cite{Chollet2024-Blog} is not yet \textit{much better} than just continuing to train the neural network for longer and doing deep learning \textit{without} search, although it is interesting that search-augmented AI models appear to be generating programs for themselves to execute in solving given tasks. Were it the case that ``deep-learning guided program search'' is not much better than more training and no search, then searching during generation should follow unfavorable logarithmic scaling laws, in the same way that models \textit{not} augmented by search do: The $\log$ of performance would improve with the $\log$ of \textit{time} spent generating, as opposed to either model or data set \textit{size}, or training time. The model could then achieve a \textit{linear} increase in performance either by \textit{exponentially} increasing its size, access to data, or training time (as in the \textcite{kaplan_scaling_2020} finding), or by exponentially increasing \textit{inference} time by doing program search. These four ``scaling laws'' resemble a classic space-time tradeoff, where one can achieve an algorithm's goals either by placing significant demands on storage space or on computation time, in each case economizing on the other, but \textit{combined}, the complexity of the algorithm remains fixed. The space-time tradeoff in question seems, unfortunately, to be of the pessimistic sort found in similar combinatorial search problems \cite{Mikko2010}. Indeed, \textcite{Jones2021} empirically finds exactly this sort of tradeoff in a similar Monte-Carlo model.

Whether or not ``deep-learning guided program search'' is currently capable of overcoming the limitations of prior sequence modeling approaches is important, as GPT-o3 recently achieved the first success at the ARC-AGI task \cite{Chollet2024-Blog}. The ARC-AGI task \cite{Chollet2019} is a puzzle whereby participants (human or AI) are given a small set of example pairs of grids of colored squares. In each pair, one grid is marked as containing an input pattern, and the other an output pattern; the task designer has constructed each output from the corresponding input using a simple rule shared across the examples, unknown to the participant. The participant is finally presented a test input, and tasked with constructing the appropriate output from the input, by guessing the rule that constructed the example outputs. The format is similar to the Raven's Progressive Matrices \cite{Raven2003}, as the examples are \textit{ad-hoc}, constructed just for the test and intended to be unfamiliar. 

Both the Raven's Progressive Matrices and ARC-AGI tasks test abstract reasoning capacity in an unfamiliar setting where participants are unlikely to have any prior knowledge, but ARC-AGI is not designed as a test of \textit{human} intelligence, as the tasks are designed to be quite easy for humans. Rather, the structure of the rules humans must guess is considerably more variable across examples than the Raven's task and cannot typically be solved by simple pattern completion, thus task serves as a model of an AI's capacity to discern structured rules from sparse examples, where the each presented rule is highly novel, that is, anything structured like it is unlikely to be in any training data, anywhere. The rules can be thought of as \textit{programs} that somehow transform the input into the output, and then the participant's goal is to find the program that most plausibly models the sparse examples provided, and crucially, \textit{to do so without prior knowledge of the puzzle at hand}. 

Although neural networks small enough to run on a laptop have been capable of solving Raven's Progressive Matrices at or above a human level for over a decade now \cite{Rasmussen2011}, ARC-AGI has proved difficult for AIs to solve anywhere near the level humans can, and only the colossal GPT-o3, running (by all appearances) for great amounts of time and at great expense, has managed to pass the bar, by using this \textit{kind} of ``program search'' Chollet refers to. If ``deep-learning guided program search'' is really a breakthrough, then AI models' capacity to solve \textit{highly structured sparse ad-hoc reasoning tasks} like ARC-AGI should progressively improve: they should eventually need orders of magnitude \textit{less} computing power to achieve the same goals as these particular methods of program search are refined. Conversely, if explicitly searching during the generation of outputs just allows us to spend time searching comparable to the data requirements of just training the neural network to perform the task at hand \textit{without} search (and either increases exponentially for a linear performance gain), then it will continue to be approximately as good to acquire more data and spend more time training our models as it will be to search through their possible future output sequences, and \textit{we will not have gained much by increasing the models' formal expressivity}.

In an interview on \textit{Machine Learning Street Talk}, \textcite{Chollet2025-StreetTalk} comments that the performance bought with the \textit{kind} of program search the o-series OpenAI models are doing is akin to a brute force approach over a well-constrained search space:
\begin{quote}
[I]f you look at the o1 model from OpenAI, you cannot attribute to it a fixed score on ARC-AGI, unless you're limiting yourself to ... a certain amount of compute. It's always possible to ... logarithmically improve your performance by throwing more compute at the problem. And of course this is true for o1, but even before that it was also true for brute force program search systems. Assuming you have the right DSL [domain-specific language], then extremely crude, basic, brute force program iteration can solve ARC at human level.
\end{quote}
He goes on to comment on the scaling laws that apply to explicit search during generation:
\begin{quote}
[T]he test time scaling law is basically this observation that, if you expend more compute, if you search further, you see a corresponding improvement in accuracy. And that relationship is logarithmic ... accuracy improves logarithmically with compute. ... this is not really new ... if you're doing brute force program search, for instance, your ability to solve a problem improves logarithmically with the amount of compute.
\end{quote}

Our approach, advanced here by further developing a Turing-complete vector-symbolic language, tends in the direction of explicitly encoding structured expressions in neural network states. The object of our approach is that traversal of the activation space of the network implies a traversal of structures, hence minimization of loss with respect to the neural network weights will train the network to produce structures on its activation space that may be interpreted as programs, if the neural network's function is designed to be sensitive to structure \cite{Fodor1988}.  

\textcite[p. 53]{Chollet2019} suggests designing a \textit{domain-specific language} (DSL) to approach tasks such as ARC-AGI. A DSL is a formal language, such as HTML or VHDL, ``tailored to a specific application domain'' \cite[p. 316]{Mernik2005}, whereby the syntax and semantics of the language primitives, as well as the language's dominant patterns, closely track the most relevant objects in the application domain, and the most relevant ways in which they can be manipulated. To a significant degree, the identification of a specific object and the manipulations proper to it are captured by the notion of a \textit{type}. Accordingly, describing the appropriate types a network's function needs to be sensitive to, and choosing efficient type encodings, such that structured representations can be evaluated efficiently, is an important part of making the use of the structures we are interested in practicable. The right type and the right encoding can collapse a degree of the complexity in treating certain kinds of objects. For instance, if we are interested in motor control, it is of particular use for a domain-specific language to express motor control primitives (as in, e.g., \cite{Brohan2023}). The integer encoding we add to the Lisp VSA is extensible beyond the narrow use-case presented here, as objects with an inductive, cyclic structure as integers have are similarly amenable to a residue encoding.

If the states of a neural network may be interpreted as expressions in a DSL according to a well-defined syntax, and the network behavior is sensitive to the objects of the DSL and their structure, then the states of the network can be said to be inherently interpretable. If the DSL is well-chosen, the basic objects decodable from the neural network's representations will be objects in the target problem domain, and the states of the network may be efficiently evaluated, which is necessary to make processing on subsequent layers of the network, sensitive as they are to structure, tractable. Further, if network is designed to use some particular encoding where the distribution of the syntax is such that that small perturbations in the activation space imply small and predictable perturbations in the structure of the representation as \textcite{Kelly2020indirect} achieved, then the structures encoded in the activation space can be made subject to gradient descent learning directly. Thus, the VSA representation agenda is: (1) to design neural networks capable of representing and learning over the right sort of structures (as, given the above discussion, we do not expect this to be possible for feedforward networks trained by backpropagation gradient descent), (2) to produce structures that can be used to efficiently compute the solutions to problems in a wide variety of domains, and (3) to encode the distributional semantics of those structures to provide a high degree of constraint on expressions with the same semantics. If the agenda is fulfilled, then, with \textit{polynomial} data, \textit{polynomial} network size, \textit{polynomial} training time, and \textit{polynomial} search time (as a function of the length of inputs), we can create models that tractably, without dependence on the continued explosion of AI computing resources, achieve a higher degree of general problem-solving capability \cite{Flanagan2025-generalintelligence} than has heretofore been possible. 

Our approach marks a step forward in representation design. Creating representations with the appropriate topology to do numeric computing efficiently, and an interpreter capable of carrying out numeric computing using our representations, we have moved closer towards structured representations that are efficiently evaluable by a neural network working in numeric, or other topologically embeddable \cite{Ozen2024} domains, such as \textit{motor representation} \cite{Mylopoulos2017, DeWolf2017}. Future work will continue improving language efficiency, embed the syntax in such a way that its distribution is learnable, and design neural networks capable of learning in a way that avails itself of structure.

\printbibliography
\end{document}